\documentclass[12pt]{article}
\usepackage[margin=1in]{geometry}                % See geometry.pdf to learn the layout options. There are lots.
\usepackage{setspace}
\usepackage{graphicx}
\usepackage{amsmath}
\usepackage{amssymb}
\usepackage{bm}
\usepackage{url}
\usepackage{epstopdf}
\usepackage{times}
\usepackage{ntheorem}
\newtheorem{thm}{Theorem}
\theorembodyfont{\upshape}
\newtheorem{remark}{Remark}
\title{Machine Learning using the Variational Predictive Information Bottleneck with a Validation Set}
\author{Sayandev Mukherjee\\ CableLabs, Sunnyvale, CA\\ \texttt{s.mukherjee@cablelabs.com}}
\begin{document}
\maketitle

\section{Abstract}
Zellner~\cite{zellner1988} modeled statistical inference in terms of information processing and postulated the \emph{Information Conservation Principle} (ICP) between the input and output of the information processing block, showing that this yielded Bayesian inference as the optimum information processing rule. Recently, Alemi~\cite{aaa2019} reviewed Zellner's work in the context of machine learning and showed that the ICP could be seen as a special case of a more general optimum information processing criterion, namely the \emph{predictive information bottleneck objective}.  However,~\cite{aaa2019} modeled machine learning as using training and test data sets only, and did not account for the use of a validation data set during training.  The present note is an attempt to extend Alemi's information processing formulation of machine learning, and the predictive information bottleneck objective for model training, to the widely-used scenario where training utilizes not only a training but also a validation data set.

\section{Review of Information Processing formulation of Machine Learning}
\subsection{Introduction and Notation}
We will use Alemi's formulation and notation from~\cite{aaa2019}, with some additional detail for clarity.  Consider a data generating process $\phi$ with distribution (PMF or PDF, depending on whether $\phi$ is discrete or continuous-valued, respectively) $p(\phi)$, which generates the features $x$ according to the distribution $p(x|\phi)$.  We collect $N$ samples of $x$ in the training set $\bm{x_P} = \{x_1,x_2,\dots,x_N\}$, with the choice of subscript `P' emphasizing that these are \emph{past} observations.  Depending on whether we are testing the performance of a trained model on a test set or deploying a trained model in a production environment to perform inference, we may have a finite or (potentially) infinite set $\bm{x_F} = \{x_{N+1},\dots,\}$ of \emph{future} (i.e., not seen during training) samples of $x$ from the same process (also emphasized by the choice of subscript `F').

\subsection{The Predictive Information Bottleneck Objective for Model Training}
Model training or ``learning" is the extraction of the model parameters $\theta$ from the training set $\bm{x_P}$.  Viewed from the perspective of information processing, we may see model training as computing, and sampling from, the distribution $p(\theta\,|\,\bm{x_P})$.  

Again from the perspective of information processing, the trained model may be evaluated by how much \emph{information} the training-derived representation $p(\theta\,|\,\bm{x_P})$ captures about future samples $\bm{x_F}$.  In other words, we want to find the $p(\theta\,|\,\bm{x_P})$ that maximizes the mutual information $I(\theta;\,\bm{x_F})$:
\[
	\max_{p(\theta\,|\,\bm{x_P})} I(\theta;\,\bm{x_F}),
\]
where for any two random variables $x$ and $y$, their mutual information $I(x;\,y)$ is defined as the Kullback-Leibler distance between their joint distribution $p(x,y)$ and the product of their marginal distributions $p(x)$ and $p(y)$:
\[
	I(x;\,y) = D_{\mathrm{KL}}\Big(p(x,y) \,\Big\|\, p(x)p(y)\Big) = \mathbb{E}_{p(x,y)} \left[\log\frac{p(x,y)}{p(x)p(y)}\right] = I(y;\,x).
\]

Alemi~\cite{aaa2019} proposes to limit the complexity of the model representation obtained from training by incorporating a \emph{bottleneck} requirement on the mutual information $I(\theta;\, \bm{x_P})$.\footnote{Note that we assume that $I(\theta;\,\bm{x_P})$ is finite and not a constant.  For this, we require that $\theta$ not be a deterministic function of $\bm{x_P}$~\cite{goldfeld2019}.  This requirement is satisfied for a deep learning model trained by gradient descent (because of random initialization) or stochastic gradient descent (because of both random initialization and randomized selection of minibatch entries by replacement from $\bm{x_P}$).}  This yields the following \emph{predictive information bottleneck objective} on training:
\begin{equation}
	 \max_{p(\theta\,|\,\bm{x_P})} I(\theta;\,\bm{x_F}) \qquad \text{ subject to } \qquad I(\theta;\, \bm{x_P}) = I_0.
	 \label{eq:a1}
\end{equation}
Applying a Lagrange multiplier $\lambda$ to the constraint on $I(\theta;\,\bm{x_P})$,~\eqref{eq:a1} can be rewritten as the unconstrained optimization problem
\begin{equation}
	\max_{\lambda, p(\theta\,|\,\bm{x_P})} [I(\theta;\,\bm{x_F}) - \lambda I(\theta;\,\bm{x_P})].
	\label{eq:a2}
\end{equation}
Note that there is no constraint on the sign of $\lambda$. Next, from the Markov property of the information processing chain $\bm{x_F} \leftarrow \phi \rightarrow \bm{x_P} \rightarrow \theta$, we have $I(\theta;\,\bm{x_F},\bm{x_P}) = I(\theta;\,\bm{x_P})$, so we have
\begin{align}
	I(\theta;\,\bm{x_F}) - I(\theta;\,\bm{x_P}) &= I(\theta;\,\bm{x_F},\bm{x_P}) - I(\theta;\,\bm{x_P}\,|\,\bm{x_F}) - I(\theta;\,\bm{x_P}) \notag \\
	&= -I(\theta;\,\bm{x_P}\,|\,\bm{x_F}), \label{eq:a3}
\end{align}
where in the first step we have used the identity
\begin{equation}
	I(x;\,y,z) = I(x;\,y) + I(x;\,z\,|\,y).
	\label{eq:id}
\end{equation}
Combining~\eqref{eq:a2} and~\eqref{eq:a3} then yields the following unconstrained optimization problem equivalent to the predictive information bottleneck objective~\eqref{eq:a1}:
\begin{equation}
	\min_{\beta, p(\theta\,|\,\bm{x_P})} [I(\theta;\,\bm{x_P}\,|\,\bm{x_F}) - \beta I(\theta;\, \bm{x_P})],
	\label{eq:a4}
\end{equation}
where $\beta = 1 - \lambda$.  Since~\eqref{eq:a4} cannot be solved directly, Alemi employs two variational approximations that are described next.

\subsection{Variational Approximations to the Predictive Information Bottleneck Objective}
\label{sec:var}
\begin{itemize}
\item Treat $\theta$ as unobserved variables and use a variational approximation $q(\theta)$ to the true distribution $p(\theta\,|\,\bm{x_F})$, where $q(\cdot)$ is a distribution chosen independent of $\bm{x_F}$.  Denoting by $\mathbb{E}$ the expectation with respect to (the true distributions of) all the random variables $\theta, \bm{x_P}, \phi, \bm{x_F}$, we have (recalling that $\theta$ and $\bm{x_F}$ are conditionally independent given $\bm{x_P}$): 
\begin{align}
	I(\theta;\,\bm{x_P}\,|\,\bm{x_F}) &= \mathbb{E}\left[\log\frac{p(\theta, \bm{x_P}\,|\,\bm{x_F})}{p(\theta\,|\,\bm{x_F}) p(\bm{x_P}\,|\,\bm{x_F})}\right] \notag \\
	&= \mathbb{E}\left[\log\frac{p(\theta\,|\,\bm{x_P}, \bm{x_F})}{p(\theta\,|\,\bm{x_F})}\right] = \mathbb{E}\left[\log\frac{p(\theta\,|\,\bm{x_P})}{p(\theta\,|\,\bm{x_F})}\right]\notag \\
	&= \mathbb{E}\left[\log\frac{p(\theta\,|\,\bm{x_P})}{q(\theta)} - \log\frac{p(\theta\,|\,\bm{x_F})}{q(\theta)}\right] \notag \\	
	&= \mathbb{E}\left[\log\frac{p(\theta\,|\,\bm{x_P})}{q(\theta)}\right] - \mathbb{E}\left\{\mathbb{E}\left[\log\frac{p(\theta\,|\,\bm{x_F})}{q(\theta)} \,\Bigg |\, \bm{x_F}\right]\right\} \notag \\
	&= \mathbb{E}\left[\log\frac{p(\theta\,|\,\bm{x_P})}{q(\theta)}\right] - \mathbb{E}\left[D_{\mathrm{KL}}\Big(p(\cdot\,|\,\bm{x_F})\,\Big\|\,q(\cdot)\Big)\right] \notag \\
	&\leq \mathbb{E}\left[\log\frac{p(\theta\,|\,\bm{x_P})}{q(\theta)}\right].  \label{eq:a5} 
\end{align}
\item Treat the distribution $p(\bm{x_P}\,|\,\theta)$ as a ``likelihood" function and use a variational approximation for it given by $q(\bm{x_P}\,|\,\theta)$, for example the factorized form $q(\bm{x_P}\,|\,\theta) = \prod_{x \in \bm{x_P}} q(x|\theta)$ for some selected distribution $q(\cdot\,|\,\theta)$.  Then we can write
\begin{align}
	I(\theta;\,\bm{x_P}) &= \mathbb{E}\left[\log\frac{p(\theta, \bm{x_P})}{p(\theta)p(\bm{x_P})}\right] = \mathbb{E}\left[\log\frac{p(\bm{x_P}\,|\,\theta)}{p(\bm{x_P})}\right] \notag \\
	&= \mathbb{E}\left[\log\frac{p(\bm{x_P}\,|\,\theta)}{q(\bm{x_P}\,|\,\theta)} -\log p(\bm{x_P}) + \log q(\bm{x_P}\,|\,\theta)\right] \notag \\
	&= \mathbb{E}\left[D_{\mathrm{KL}}\Big(p(\cdot\,|\,\theta)\,\Big\|\,q(\cdot\,|\,\theta)\Big)\right] +\mathbb{E}[-\log p(\bm{x_P})] + \mathbb{E}[\log q(\bm{x_P}\,|\,\theta)] \notag \\
	&\geq 0 + H(\bm{x_P}) + \mathbb{E}[\log q(\bm{x_P}\,|\,\theta)], \label{eq:a6}
\end{align}
where for any random variable $x$, $H(x)$ is its \emph{entropy}:
\[
	H(x) = -\sum_x p(x) \log p(x) \qquad \text{ or } \qquad H(x) = -\int p(x) \log p(x)\,\mathrm{d}x.
\]
\end{itemize}

\subsection{Variational Formulation of Predictive Information Bottleneck Objective}
\label{sec:varpibo}
From~\eqref{eq:a5} and~\eqref{eq:a6} we therefore have, for every choice of $\beta \geq 0$, variational approximate marginal distribution $q(\theta)$, and variational approximate likelihood function $q(x\,|\,\theta)$, the following upper bound on the objective function of~\eqref{eq:a4}:
\begin{equation}
	I(\theta;\,\bm{x_P}\,|\,\bm{x_F}) - \beta I(\theta;\,\bm{x_P}) \leq \mathbb{E}\left[\log\frac{p(\theta\,|\,\bm{x_P})}{q(\theta)}\right] - \beta\,\mathbb{E}[\log q(\bm{x_P}\,|\,\theta)] - \beta H(\bm{x_P}).
	\label{eq:a7}
\end{equation}

Note that $H(\bm{x_P})$ is a constant outside our control.  Since the exact problem~\eqref{eq:a4} cannot be solved directly, we can simply select $\beta \geq 0, q(\theta), q(x\,|\,\theta)$ based on some external criteria and solve the following problem, whose optimum value yields an upper bound on the exact objective in~\eqref{eq:a4}:
\begin{align}
	& \min_{p(\theta\,|\,\bm{x_P})} \mathbb{E}_{p(\phi)p(\bm{x_P}\,|\,\phi)p(\theta\,|\,\bm{x_P})}\left[\log\frac{p(\theta\,|\,\bm{x_P})}{q(\theta)} - \beta \log q(\bm{x_P}\,|\,\theta)\right] \label{eq:a8m} \\
	&= \min_{p(\theta\,|\,\bm{x_P})} \mathbb{E}_{p(\phi)p(\bm{x_P}\,|\,\phi)}\left\{\mathbb{E}_{p(\theta\,|\,\bm{x_P})}\left[\log\frac{p(\theta\,|\,\bm{x_P})}{q(\theta) q(\bm{x_P}\,|\,\theta)^\beta}\,\bigg|\,\bm{x_P}\right]\right\} \notag \\
	&= \min_{p(\theta\,|\,\bm{x_P})} \mathbb{E}_{p(\phi)p(\bm{x_P}\,|\,\phi)}\left[D_{\mathrm{KL}}\Bigg(p(\cdot\,|\,\bm{x_P})\,\Bigg\|\,\frac{q(\cdot)q(\bm{x_P}\,|\,\cdot)^\beta}{Z_\beta(\bm{x_P})}\Bigg)\right],
	\label{eq:a8}
\end{align}
where 
\[
	Z_\beta(\bm{x_P}) = \sum_{\psi}q(\psi)q(\bm{x_P}\,|\,\psi)^\beta \qquad \text{ or } \qquad 	Z_\beta(\bm{x_P}) = \int q(\psi)q(\bm{x_P}\,|\,\psi)^\beta \,\mathrm{d}\psi
\]
and we emphasize that the expectation in~\eqref{eq:a8m} is with respect to the true distributions of $\theta,\bm{x_P},\phi$.  Note that the use of the variational approximation $q(\theta)$ has eliminated the dependence on the distribution of $\bm{x_F}$.

Finally, it follows from~\eqref{eq:a8} that the optimum distribution $p(\theta\,|\,\bm{x_P})$ is given by
\[
	p(\theta\,|\,\bm{x_P}) = \frac{q(\theta) \, q(\bm{x_P}\,|\,\theta)^\beta}{Z_\beta(\bm{x_P})}.
\]

For $\beta=1$, the objective in~\eqref{eq:a8} can be identified as the ICP postulated by Zellner~\cite{zellner1988}, and the optimum $p(\theta\,|\,\bm{x_P})$ is the Bayesian inference derived from the variational marginal and likelihood $q(\theta)$ and $q(x\,|\,\theta)$ respectively:
\[
	p(\theta\,|\,\bm{x_P}) \propto q(\theta) \, q(\bm{x_P}\,|\,\theta).
\]

\section{The Modified Predictive Information Bottleneck Objective}
\subsection{Shortcomings of the Predictive Information Bottleneck Objective~\eqref{eq:a1}}
Alemi's formulation~\cite{aaa2019} of the predictive information bottleneck objective in~\eqref{eq:a1} has the following shortcomings:
\begin{itemize}
\item Recall that in~\eqref{eq:a1}, the Lagrange multiplier $\lambda$ may be positive or negative-valued.  However, the variational upper bound in~\eqref{eq:a7} does not hold unless $\beta \geq 0$, i.e., $\lambda \leq 1$, though there is no explanation offered in~\cite{aaa2019} of why only values of $\lambda$ in this range make sense for this problem.
\item The quantity $I_0$ is never used after the equality constraint $I(\theta;\,\bm{x_P}) = I_0$ is incorporated into the problem formulation in~\eqref{eq:a1}.  Further, $\beta$ is simply selected as per other criteria, and not set so as to achieve this equality $I(\theta;\,\bm{x_P}) = I_0$.
\end{itemize}

\subsection{Predictive Information Bottleneck Objective with Inequality Constraint}
\label{sec:ineq}
The fundamental reason for the above shortcomings is the requirement of \emph{equality} in the bottleneck constraint.

We therefore propose to modify the predictive information bottleneck objective from~\eqref{eq:a1} to one that has an inequality constraint between $I(\theta;\,\bm{x_P})$, the mutual information of the trained model and the training set, and a pre-selected threshold $I_0$:  
\begin{equation}
	\max_{p(\theta\,|\,\bm{x_P})} I(\theta;\,\bm{x_F}) \qquad \text{ subject to } \qquad I(\theta;\,\bm{x_P}) \geq I_0 \Leftrightarrow I_0 - I(\theta;\,\bm{x_P}) \leq 0.
	\label{eq:b1}
\end{equation}
Note that this formulation imposes a bottleneck on model performance, such that only models that extract a certain threshold level of information from the training set clear the bottleneck.
From the Karush-Kuhn-Tucker (KKT) theorem, the optimization problem~\eqref{eq:a1} now changes to
\begin{align}
	\max_{\mu \geq 0, p(\theta\,|\,\bm{x_P})} \left\{I(\theta;\,\bm{x_F}) - \mu\Big[I_0 - I(\theta;\,\bm{x_P})\Big]\right\} &= \max_{\mu \geq 0, p(\theta\,|\,\bm{x_P})} \Big[I(\theta;\,\bm{x_F}) + \mu I(\theta;\,\bm{x_P})\Big] \label{eq:b2} \\
	&= \min_{\beta \geq 1, p(\theta\,|\,\bm{x_P})} \Big[I(\theta;\,\bm{x_P}\,|\,\bm{x_F}) - \beta I(\theta;\,\bm{x_P})\Big],
	\label{eq:b3}
\end{align}
where we have used~\eqref{eq:a3} to go from~\eqref{eq:b2} to~\eqref{eq:b3}, and $\beta = 1 + \mu \geq 1$ because $\mu \geq 0$ by definition.   Note that this immediately resolves the interpretability issues\footnote{Except for the uninteresting case where $\beta \downarrow 0$, i.e., we ignore the data altogether while training (see~\cite[Table~1]{aaa2019}).} that arose with the equality bottleneck constraint formulation above.  Note also that the objective function of~\eqref{eq:b3} is the same as the left hand side of~\eqref{eq:a7}.  Then the variational approximations discussed in Sec.~\ref{sec:var} can be applied as before, resulting in the same variational objective function~\eqref{eq:a8}. 

Further, the KKT theorem requires that the optimum $\mu, p(\theta\,|\,\bm{x_P})$ for~\eqref{eq:b2} satisfy the \emph{complementary slackness} condition
\begin{equation}
	\mu \Big[I_0 - I(\theta;\,\bm{x_P})\Big] = 0,
	\label{eq:b4}
\end{equation}
which implies that if the training step achieves $I(\theta;\,\bm{x_P}) > I_0$, then the optimum $\mu$ is zero, i.e., the optimum $\beta$ in~\eqref{eq:b3} is unity.  From the discussion at the end of Sec.~\ref{sec:varpibo} we have the following result. 

\begin{thm} \label{thm:1} If training step is successful, i.e., we achieve $I(\theta;\,\bm{x_P}) > I_0$, then the optimum information processing for~\eqref{eq:a8m} is Bayesian inference:
\begin{equation}
	p(\theta\,|\,\bm{x_P}) \propto q(\theta) \, q(\bm{x_P}\,|\,\theta),
	\label{eq:b5}
\end{equation}
where $q(\theta)$ and $q(x\,|\,\theta)$ are as defined in Sec.~\ref{sec:var}.
\end{thm}

\begin{remark} \label{remark:1} In a practical scenario where we train, say, a chosen deep learning architecture with stochastic gradient descent over mini-batches sampled from the training set, ``success" in training is defined as some measure of performance on the training set, such as average training loss (if the model is being trained for a regression or function approximation application) or classification error (if the model is being trained for classification), exceeding some threshold. In the present information-theoretic framework, this performance threshold will need to be translated into some threshold $I_0$ on the mutual information $I(\theta;\,\bm{x_P})$.  Now, $I(\theta;\,\bm{x_P})$ cannot be computed or even estimated directly, but we can use the variational lower bound~\eqref{eq:a6} on $I(\theta;\,\bm{x_P})$ as follows:
\begin{itemize}
\item Estimate the entropy $H(\bm{x_P})$ of the training set using a source-coding or compression scheme;
\item Estimate $L \equiv \mathbb{E}[\log q(\bm{x_P}\,|\,\theta)]$ from an average over mini-batches sampled from $\bm{x_P}$;
\item Declare ``successful training" when $\hat{H}(\bm{x_P}) + \hat{L} > I_0$, where the hats denote the estimates.
\end{itemize}
\end{remark}

\section{Extension to Model Training with a Validation Set}
\subsection{Introduction and Notation}
The machine learning model is trained on a training set $\bm{x_P}$.  For training classical machine learning models like support vector machines, decision trees, and random forests, which are not as computationally expensive to train as deep learning models, this training set $\bm{x_P}$ can be used for \emph{cross-validation} to validate the performance of the model during training in order to guard against overfitting the training set. However, $k$-fold cross-validation requires the model to be trained from scratch $k$ times, which often imposes an unacceptable computational burden when the model is a deep learning one.  Thus, when training deep learning models, it is widespread practice to set aside a portion of the available training data as a \emph{validation set}, use the remainder of the training data as the \emph{training set} to train the model, and, as the model trains, validate its performance on the validation set.

In the notation of the previous section, if the available training data comprises the $N$ samples $x_1,x_2,\dots,x_N$, the training set is now $\bm{x_P} = \{x_1,x_2,\dots,x_M\}$ where $M < N$, with the rest of the training data comprising the validation set $\bm{x_V} = \{x_{M+1},\dots,x_N\}$.  

\subsection{Predictive Information Bottleneck Objective with Validation Set}
As in Sec.~\ref{sec:ineq}, we will require the trained model to clear the bottleneck of a threshold on mutual information between the model and the training set: 
\begin{equation}
	I(\theta\,;\,\bm{x_P}) \geq I_0'.
	\label{eq:v1}
\end{equation}  
In addition, we will require the model, after being trained on the training set $\bm{x_P}$, to clear another threshold of mutual information between the model and the validation set (analogous to the performance requirement on mutual information between the model and the test set $\bm{x_F}$):
\begin{equation}
	I(\theta\,;\,\bm{x_V}\,|\,\bm{x_P}) \geq I_0''.
	\label{eq:v2}
\end{equation}
From~\eqref{eq:id} we see that~\eqref{eq:v1} and~\eqref{eq:v2} together yield
\begin{equation}
	I(\theta\,;\,\bm{x_P},\bm{x_V}) \geq I_0 \equiv I_0' + I_0'',
	\label{eq:v3}
\end{equation}
which is equivalent to requiring the bottleneck condition with threshold $I_0 = I_0' + I_0''$ on the mutual information between the model and the augmented training set $\{x_1,x_2,\dots,x_N\}$ created from $\bm{x_P}$ and $\bm{x_V}$.  However, instead of requiring the single condition~\eqref{eq:v3}, we shall impose the stricter requirement to satisfy the two conditions~\eqref{eq:v1} and~\eqref{eq:v2}. 

The new optimization problem is now
\begin{equation}
	\max_{p(\theta\,|\,\bm{x_P},\bm{x_V})} I(\theta;\,\bm{x_F}) \qquad \text{ subject to } \qquad I(\theta;\,\bm{x_P}) \geq I_0' \quad \text{ and } \quad I(\theta;\,\bm{x_V}\,|\,\bm{x_P}) \geq I_0''.
	\label{eq:v4}
\end{equation}
In~\eqref{eq:v4} the distribution $p(\theta\,|\,\bm{x_P},\bm{x_V})$ incorporates the dependence of the trained model on the validation set.  This dependence arises because we train by running a learning algorithm on (a subset of) the training set,  validate the (partially) trained model on the validation set, run the learning algorithm again on (another subset of) the training set, validate the (partially) trained model again on the validation set, and so on until the performance of the model on both the training and validation sets satisfies some criteria which we represent here by~\eqref{eq:v1} and~\eqref{eq:v2} respectively.

\subsection{Modified Predictive Information Bottleneck Objective with Inequality Constraint and Validation Set}
From the Markov property of the information processing chain $\bm{x_F} \leftarrow \phi \rightarrow (\bm{x_P},\bm{x_V}) \rightarrow \theta$, we have $I(\theta;\,\bm{x_F},(\bm{x_P},\bm{x_V})) = I(\theta;\,\bm{x_P},\bm{x_V})$, so we have
\begin{align}
	& I(\theta;\,\bm{x_F}) - I(\theta;\,\bm{x_P}) - I(\theta;\,\bm{x_V}\,|\,\bm{x_P}) \notag \\
	&= I(\theta;\,\bm{x_F}) - I(\theta;\,\bm{x_P},\bm{x_V}) \notag \\
	&= I(\theta;\,\bm{x_F},(\bm{x_P},\bm{x_V})) - I(\theta;\,\bm{x_P},\bm{x_V}\,|\,\bm{x_F}) - I(\theta;\,\bm{x_P},\bm{x_V}) \qquad [\text{from~\eqref{eq:id}}]\notag \\
	&= -I(\theta\,;\, \bm{x_P}, \bm{x_V} \,|\,\bm{x_F}). \label{eq:v5}
\end{align}

As with~\eqref{eq:b1}, the KKT theorem lets us rewrite~\eqref{eq:v4} as follows:
\begin{align}
	& \max_{\mu \geq 0, \nu \geq 0, p(\theta\,|\,\bm{x_P},\bm{x_V})} \left\{I(\theta;\,\bm{x_F}) - \mu\Big[I_0' - I(\theta;\,\bm{x_P})\Big] - \nu\Big[I_0'' - I(\theta;\,\bm{x_V}\,|\,\bm{x_P})\Big]\right\} \notag \\
	&= \max_{\mu \geq 0, \nu \geq 0, p(\theta\,|\,\bm{x_P},\bm{x_V})} \Big[I(\theta;\,\bm{x_F}) + \mu I(\theta;\,\bm{x_P}) + \nu I(\theta;\,\bm{x_V}\,|\,\bm{x_P})\Big] \label{eq:v6} \\
	&= \min_{\beta \geq 1, \gamma \geq 1, p(\theta\,|\,\bm{x_P},\bm{x_V})} \Big[I(\theta\,;\, \bm{x_P}, \bm{x_V} \,|\,\bm{x_F}) - \beta I(\theta;\,\bm{x_P}) - \gamma I(\theta;\,\bm{x_V}\,|\,\bm{x_P})\Big], \label{eq:v7}
\end{align}
where we use~\eqref{eq:v5} to go from~\eqref{eq:v6} to~\eqref{eq:v7}, and $\beta = 1 + \mu \geq 1$, $\gamma = 1 + \nu \geq 1$.  Further, the optimum $\beta, \gamma, p(\theta\,|\,\bm{x_P},\bm{x_V})$ must satisfy the complementary slackness conditions
\begin{align}
	\mu\Big[I_0' - I(\theta;\,\bm{x_P})\Big] &= 0, \label{eq:v7a} \\
	\nu\Big[I_0'' - I(\theta;\,\bm{x_V}\,|\,\bm{x_P})\Big] &= 0. \label{eq:v7b}
\end{align}
	 
\subsection{Variational Approximations}
\label{sec:var_pibo_ineq}
\begin{itemize}
\item Treat $\theta$ as unobserved variables and use a variational approximation $q(\theta)$ to the true distribution $p(\theta\,|\,\bm{x_F})$, where $q(\cdot)$ is a distribution chosen independent of $\bm{x_F}$.  Following the notation of Sec.~\ref{sec:var} and the same steps as in the derivation of~\eqref{eq:a5}, we obtain
\begin{equation}
	I(\theta\,;\,\bm{x_P},\bm{x_V}\,|\,\bm{x_F}) \leq \mathbb{E}\left[\log\frac{p(\theta\,|\,\bm{x_P},\bm{x_V})}{q(\theta)}\right].
	\label{eq:v10}
\end{equation}
\item Treat the distribution $p(\bm{x_P}\,|\,\theta)$ as a ``likelihood" function and use a variational approximation for it given by $q(\bm{x_P}\,|\,\theta)$, for example the factorized form $q(\bm{x_P}\,|\,\theta) = \prod_{x \in \bm{x_P}} q(x|\theta)$ for some selected distribution $q(\cdot\,|\,\theta)$.  Then we obtain~\eqref{eq:a6}:
\[
	I(\theta;\,\bm{x_P}) \geq H(\bm{x_P}) + \mathbb{E}[\log q(\bm{x_P}\,|\,\theta)].
\]	
\item Treat the conditional distribution $p(\bm{x_V}\,|\,\bm{x_P},\theta)$ as a conditional likelihood function and use a variational approximation for it given by $q(\bm{x_V}\,|\,\bm{x_P},\theta)$, for example the factorized form $q(\bm{x_V}\,|\,\bm{x_P},\theta) = \prod_{x \in \bm{x_V}} q(x\,|\,\bm{x_P},\theta)$ for some selected distribution $q(\cdot\,|\,\bm{x_P}, \theta)$.  Then we obtain
\begin{align}
	& I(\theta;\,\bm{x_V}\,|\,\bm{x_P}) \notag \\
	&= \mathbb{E}\left[\log\frac{p(\theta, \bm{x_V}\,|\,\bm{x_P})}{p(\theta\,|\,\bm{x_P})p(\bm{x_V}\,|\,\bm{x_P})}\right] = \mathbb{E}\left[\log\frac{p(\theta, \bm{x_V}\,|\,\bm{x_P})}{p(\theta\,|\,\bm{x_P})p(\bm{x_V}\,|\,\bm{x_P})}\right] \notag \\
	&= \mathbb{E}\left[\log\frac{p(\bm{x_V}\,|\,\bm{x_P},\theta)}{p(\bm{x_V}\,|\,\bm{x_P})}\right] \notag \\
	&= \mathbb{E}\left[\log\frac{p(\bm{x_V}\,|\,\bm{x_P},\theta)}{q(\bm{x_V}\,|\,\bm{x_P},\theta)} -\log p(\bm{x_V}\,|\,\bm{x_V}) + \log q(\bm{x_V}\,|\,\bm{x_P},\theta)\right] \notag \\
	&= \mathbb{E}\left[D_{\mathrm{KL}}\Big(p(\cdot\,|\,\bm{x_P},\theta)\,\Big\|\,q(\cdot\,|\,\bm{x_P},\theta)\Big)\right] +\mathbb{E}\left[-\log p(\bm{x_V}\,|\,\bm{x_P})\right] + \mathbb{E}[\log q(\bm{x_V}\,|\,\bm{x_P},\theta)] \notag \\
	&\geq 0 + H(\bm{x_V}\,|\,\bm{x_P}) + \mathbb{E}[\log q(\bm{x_V}\,|\,\bm{x_P},\theta)], 
	\label{eq:v11}
\end{align}
where for any two random variables $x$ and $y$,
\[
	H(x\,|\,y) = -\sum_{x,y} p(x,y) \log p(x\,|\,y) \qquad \text{ or } \qquad H(x\,|\,y) = -\iint p(x,y)\log p(x\,|\,y) \,\mathrm{d}x\,\mathrm{d}y.
\]

Note that the process of training on $\bm{x_P}$ along with periodic validation using the validation set $\bm{x_V}$ induces a dependence between $\bm{x_V}$ and $\bm{x_P}$ through the model $\theta$, hence it is not true in general that $p(\bm{x_V}\,|\,\bm{x_P},\theta) = p(\bm{x_V}\,|\,\theta)$ even if $\bm{x_V}$ and $\bm{x_P}$ are assumed independent.
\end{itemize}

\subsection{Variational Formulation of the Predictive Information Bottleneck \\Objective with Inequality Constraint and a Validation Set}
From~\eqref{eq:v10},~\eqref{eq:a6}, and~\eqref{eq:v11}, the objective function in~\eqref{eq:v7} is upper-bounded for each choice of $\beta \geq 1, \gamma \geq 1, q(\theta), q(x\,|\,\theta), q(x\,|\,\bm{x_P},\theta)$ as follows:
\begin{align}
	& I(\theta;\,\bm{x_P}, \bm{x_V} \,|\,\bm{x_F}) - \beta I(\theta;\,\bm{x_P}) - \gamma I(\theta;\,\bm{x_V}\,|\,\bm{x_P}) \notag \\
	&\leq \mathbb{E}\left[\log\frac{p(\theta\,|\,\bm{x_P},\bm{x_V})}{q(\theta)} - \beta \log q(\bm{x_P}\,|\,\theta) -\gamma \log q(\bm{x_V}\,|\,\bm{x_P},\theta)\right] - \beta H(\bm{x_P}) - \gamma H(\bm{x_V}\,|\,\bm{x_P}) \notag \\
	&= \mathbb{E}\left[\log\frac{p(\theta\,|\,\bm{x_P},\bm{x_V})}{q(\theta)} - \beta \log q(\bm{x_P}\,|\,\theta) -\gamma \log q(\bm{x_V}\,|\,\bm{x_P},\theta)\right] \notag \\
	&\qquad - \beta H(\bm{x_P}) - \gamma H(\bm{x_V}) + \gamma I(\bm{x_P};\,\bm{x_V}),
	\label{eq:v12}
\end{align}
where in the final step we use the identity that for any two random variables $x$ and $y$,
\[
	I(x;\,y) = H(x) - H(x\,|\,y).
\]

We treat the last three terms on the right hand side of~\eqref{eq:v12} as constants outside our control.  Since the exact problem~\eqref{eq:v7} cannot be solved directly, we simply select $\beta \geq 1$, $\gamma \geq 1$, $q(\theta)$, $q(x\,|\,\theta)$, and $q(x\,|\,\bm{x_P},\theta)$ based on some external criteria and solve the following problem, whose optimum value yields an upper bound on the exact objective in~\eqref{eq:v7}:
\begin{align}
	& \min_{p(\theta\,|\,\bm{x_P},\bm{x_V})} \mathbb{E}_{p(\phi)p(\bm{x_P},\bm{x_V}\,|\,\phi)p(\theta\,|\,\bm{x_P},\bm{x_V})}\left[\log\frac{p(\theta\,|\,\bm{x_P},\bm{x_V})}{q(\theta)} - \beta \log q(\bm{x_P}\,|\,\theta) -\gamma \log q(\bm{x_V}\,|\,\bm{x_P},\theta)\right] \label{eq:v13m} \\
	&= \min_{p(\theta\,|\,\bm{x_P},\bm{x_V})} \mathbb{E}_{p(\phi)p(\bm{x_P},\bm{x_V}\,|\,\phi)}\left\{\mathbb{E}_{p(\theta\,|\,\bm{x_P},\bm{x_V})}\left[\log\frac{p(\theta\,|\,\bm{x_P},\bm{x_V})}{q(\theta)q(\bm{x_P}\,|\,\theta)^\beta q(\bm{x_V}\,|\,\bm{x_P},\theta)^\gamma} \,\bigg|\,\bm{x_P},\bm{x_V}\right]\right\} \notag \\
	&= \min_{p(\theta\,|\,\bm{x_P},\bm{x_V})} \mathbb{E}_{p(\phi)p(\bm{x_P},\bm{x_V}\,|\,\phi)}\left[D_{\mathrm{KL}}\Bigg(p(\cdot\,|\,\bm{x_P},\bm{x_V})\,\Bigg\|\,\frac{q(\cdot)q(\bm{x_P}\,|\,\cdot)^\beta q(\bm{x_V}\,|\,\bm{x_P},\cdot)^\gamma}{Z_{\beta,\gamma}(\bm{x_P},\bm{x_V})}\Bigg)\right],
	\label{eq:v13}
\end{align}
where 
\begin{align*}
	Z_{\beta,\gamma}(\bm{x_P},\bm{x_V}) &= \sum_{\psi} q(\psi)q(\bm{x_P}\,|\,\psi)^\beta q(\bm{x_V}\,|\,\bm{x_P},\psi)^\gamma \\
	& \text{ or } \\ 
	Z_{\beta,\gamma}(\bm{x_P},\bm{x_V}) &= \int q(\psi)q(\bm{x_P}\,|\,\psi)^\beta q(\bm{x_V}\,|\,\bm{x_P},\psi)^\gamma \,\mathrm{d}\psi
\end{align*}
and we emphasize that the expectation in~\eqref{eq:v13m} is over the true distributions of $\theta$, $(\bm{x_P},\bm{x_V})$, $\phi$.  Note that the use of the variational approximation $q(\theta)$ has eliminated the dependence on the distribution of $\bm{x_F}$.

It follows from~\eqref{eq:v13} that the optimum distribution $p(\theta\,|\,\bm{x_P},\bm{x_V})$ is given by
\[
	p(\theta\,|\,\bm{x_P},\bm{x_V}) = \frac{q(\theta) q(\bm{x_P}\,|\,\theta)^\beta q(\bm{x_V}\,|\,\bm{x_P},\theta)^\gamma}{Z_{\beta,\gamma}(\bm{x_P},\bm{x_V})}.
\]

Further, from~\eqref{eq:v7a} and~\eqref{eq:v7b}, we can prove the following result in the same way as Theorem~\ref{thm:1}.
\begin{thm} \label{thm:2} If training with the validation set is successful, i.e., we achieve $I(\theta;\,\bm{x_P}) > I_0'$ and $I(\theta;\,\bm{x_V}\,|\,\bm{x_P}) > I_0''$, then the optimum information processing for~\eqref{eq:v13m} is Bayesian inference:
\begin{equation}
	p(\theta\,|\,\bm{x_P}) \propto q(\theta) \, q(\bm{x_P}\,|\,\theta) \, q(\bm{x_V}\,|\,\bm{x_P},\theta),
	\label{eq:v14}
\end{equation}
where $q(\theta)$, $q(\bm{x_V}\,|\,\theta)$, and $q(\bm{x_V}\,|\,\bm{x_P},\theta)$ are as defined in Sec.~\ref{sec:var_pibo_ineq}.
\end{thm}

\end{document}